\documentclass[letterpaper, 11pt, conference]{ieeeconf}      % Use this line for a4 paper
\IEEEoverridecommandlockouts                              % This command is only needed if 
\usepackage[left=2.3cm,top=2.3cm,right=2.3cm,bottom=2.3cm]{geometry}
\usepackage{graphicx, wrapfig, xcolor}
\usepackage{epsfig} % for postscript graphics files
\usepackage{amsmath} % assumes amsmath package installed
\usepackage{amssymb}  % assumes amsmath package installed
\usepackage{hyperref}

\usepackage{authblk}

\usepackage{dblfloatfix}    % To enable figures at the bottom of page

\title{\LARGE \bf
Accelerated Convolutions for Efficient Multi-Scale \\Time to Contact Computation in Julia}

\author{Alexander Amini, Berthold Horn, Alan Edelman\\\\Department of Electrical Engineering and Computer Science\\Massachusetts Institute of Technology, Cambridge, MA}

\begin{document}

\newcount\colveccount
\newcommand*\colvec[1]{
        \global\colveccount#1
        \begin{bmatrix}
        \colvecnext
}
\def\colvecnext#1{
    #1
    \global\advance\colveccount-1
    \ifnum\colveccount>0
            \\
            \expandafter\colvecnext
    \else
            \end{bmatrix}
    \fi
}

\maketitle
\thispagestyle{empty}
\pagestyle{empty}

%%%%%%%%%%%%%%%%%%%%%%%%%%%%%%%%%%%%%%%%%%%%%%%%%%%%%%%%%%%%%%%%%%%%%%%%%%%%%%%%
\begin{abstract}
Convolutions have long been regarded as fundamental to applied mathematics, physics and engineering. Their mathematical elegance allows for common tasks such as numerical differentiation to be computed efficiently on large data sets. Efficient computation of convolutions is critical to artificial intelligence in real-time applications, like machine vision, where convolutions must be continuously and efficiently computed on tens to hundreds of kilobytes per second. In this paper, we explore how convolutions are used in fundamental machine vision applications. We present an accelerated n-dimensional convolution package in the high performance computing language, Julia, and demonstrate its efficacy in solving the time to contact problem for machine vision. Results are measured against synthetically generated videos and quantitatively assessed according to their mean squared error from the ground truth. We achieve over an order of magnitude decrease in compute time and allocated memory for comparable machine vision applications. All code is packaged and integrated into the official Julia Package Manager to be used in various other scenarios. 

\end{abstract}

%%%%%%%%%%%%%%%%%%%%%%%%%%%%%%%%%%%%%%%%%%%%%%%%%%%%%%%%%%%%%%%%%%%%%%%%%%%%%%%%

\section{Introduction}

One of the most important concepts in machine vision \cite{ttc}, signal processing \cite{signals}, and Fourier theory \cite{fourier} is that of convolutions. Convolutions are used to compute discrete derivatives in a variety of industrial applications, such as computer graphics and geometry processing.  Additionally, with the advent of deep learning, convolutions have gained a resurgence of prominence, as Convolutional Neural Networks (CNNs) are used to compute optimized feature maps. Thus, because of their wide applicability, mathematical elegance, and fundamental symmetry between time and frequency domains, convolutions are an essential part of any mathematical computing platform.  

Two critical machine vision computations are time to contact (TTC) and focus of expansion (FOE). Imagine an autonomous vehicle, with a camera mounted on its front, approaching a wall. The time to contact (TTC) is defined as the amount of time that would elapse before the optical center reaches the surface being viewed \cite{ttc}. This problem can intuitively be thought of as: \textit{how much time will pass before the car collides with the wall?} On the other hand, the focus of expansion (FOE) will determine the precise location on the image plane that the camera is approaching (i.e., the point that would ultimately collide first). TTC and FOE solutions are critical for many robotic systems since they provide a rough safety control capability, based on continuously avoiding collision with objects around it.

The problem of accurately determining the TTC is particularly difficult because we are often not given any information of speed, or size of objects in our environment. While the FOE can be coarsely estimated by looking for minima of the directional derivatives from the video sequence, this would not give us any information relating to velocity (and therefore, the TTC). 

In the following sections, this paper will outline the formulation of discrete derivatives and convolutions, specifically in the context of machine vision and time to contact.  We then present our accelerated convolution approach (FastConv) and its implementation in Julia, a dynamic programming language for high performance computing environments \cite{julia}.  We evaluate the performance of FastConv against the existing Julia machine vision toolset and other state-of-the-art tools.  We further assess FastConv specifically for the TTC problem by implementing a multi-scale approach to TTC, and evaluate the performance against synthetically generated video data. We show that the convolution operation can be accelerated by an order of magnitude to decrease compute time and allocated memory for comparable machine vision applications over 10 times. 

\section{Discrete Derivatives}

\begin{figure*}[b!]
\centering
\includegraphics[width=0.9\linewidth]{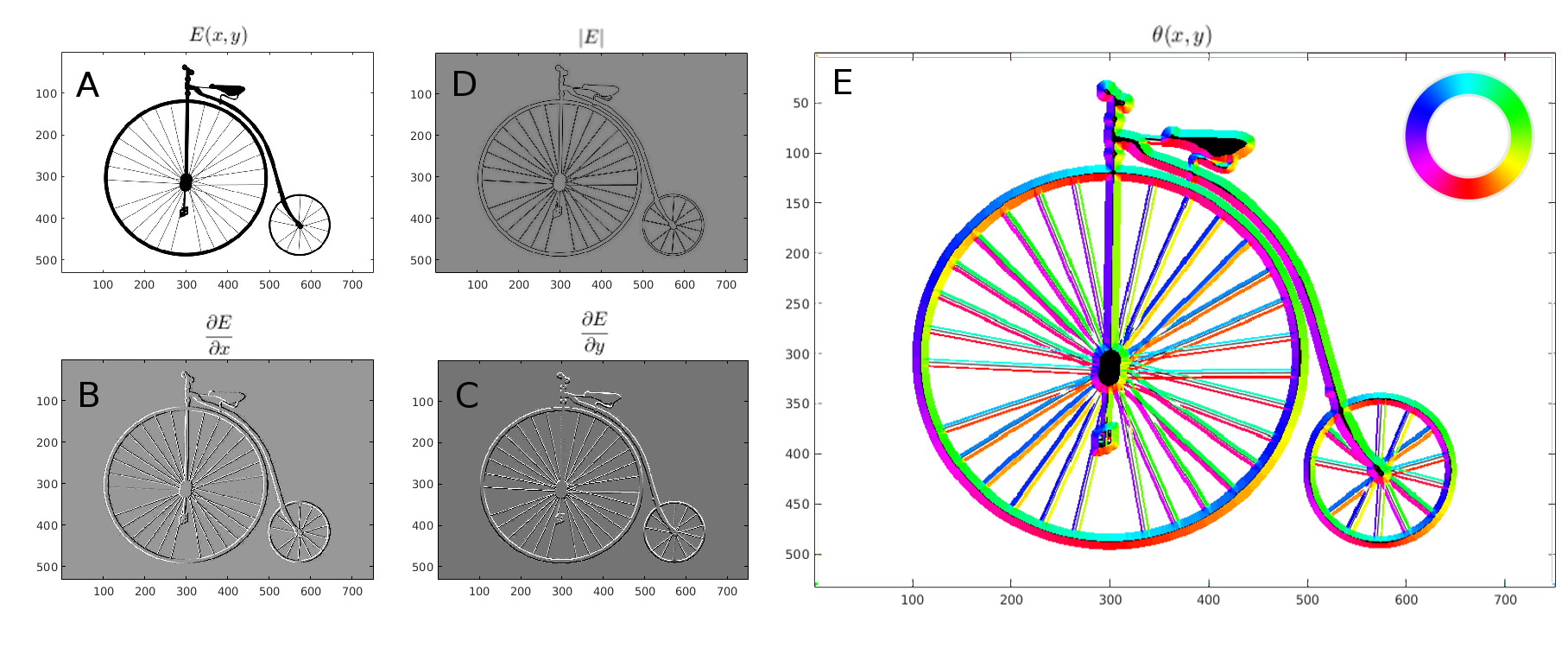}
\caption{Gradient computations of a single image $E(x,y)$ (A). Directional partial derivatives are computed along the $x$ (B) and $y$ (C) axes, along with the absolute magnitude (D). Directions of the gradient are visualized according to a circular colorbar ranging from $-\pi$ to $\pi$ (top right). } 
\label{fig:gradients}
\end{figure*}

Most fundamental machine vision techniques such as time to contact \cite{ttc}, optical flow \cite{flow}, and view synthesis \cite{viewsyn} perform several convolutions on every iteration. In the context of robotics, these techniques are performed on a continuous video stream sampled somewhere around the range of $10-30$ Hz. Since these computations need to be done at such a high frequency, it is imperative for the programming language to perform these convolutions extremely quickly. 

A fundamental part of solving the problem of TTC is computing the partial derivatives $E_x$, $E_y$, and $E_t$. Since our images are sampled under a discrete interval (determined by the number of pixels in each dimension), it is impossible for us to compute these as continuous derivatives. Instead, we need a way to \textit{estimate} the derivative in each of these directions. 

Consider a 2-dimensional (2D) image $E(x,y)$ with a single vertical edge. Since an edge is an abrupt change in brightness intensity it is the most common approach to determining meaningful discontinuities. We would expect a large magnitude impulse at the $x$ location of the edge when computing the partial derivative with respect to $x$ (ie. $E_x$). 

We state the partial derivative of an image $E(x,y)$ with respect to $x$ as: $\frac{\partial E}{\partial x}=E_x$. Similarly the derivative with respect to $y$ can be written as: $\frac{\partial E}{\partial y}=E_y$. Subsequently, we define the magnitude and angle of the edge as:

\begin{equation}
| E | = E_{mag} = \sqrt{E_x^2 + E_y^2}
\label{eq:Emag}
\end{equation}
\begin{equation}
\theta = E_{dir} = \tan^{-1}\left(\frac{E_y}{E_x}\right)
\label{eq:Eangle}
\end{equation}

Here, $| E |$ represents the sum of squares magnitude of the gradient at every pixel, and $\theta$ is the corresponding direction of the gradient. The direction is estimated by using four quadrant inverse tangent with the two directional derivatives. Note that $\theta \in [-\pi, \pi]$. Figure~\ref{fig:gradients} gives an example of taking a single image (eg. of a bicycle) and computing the resulting partial derivatives, $E_x$ (\ref{fig:gradients}B) and $E_y$ (\ref{fig:gradients}C). According to equation (\ref{eq:Emag}) the magnitude is computed in \ref{fig:gradients}D. Finally, the directions of the derivatives are computed and visualized in \ref{fig:gradients}E, where the color represents the direction (equation~\ref{eq:Eangle}) of the edge at that point (following the color wheel on the top right corner). For example, a pixel colored red indicates a gradient at that location with angle of $\frac{-\pi}{2}$.

These partial derivatives are computed according to the limit definition of a derivative: 

\[
\frac{\partial E(x,y)}{\partial x} = \lim_{\Delta x \to 0} \frac{E(x+\Delta x,y)-E(x,y)}{\Delta x}
\]

For images, a crude estimate would be to say that $\min \Delta x = 1$ since the smallest unit of measure in our image, without going to the sub-pixel scale, is a single pixel. Which in turn, would allow us to conclude that the derivative estimate in the $x$ direction for our image could be approximated by: 

\[
\frac{\partial E(x,y)}{\partial x} = E(x+1,y)-E(x,y)
\]

This operation amounts to applying a two cell mask  $k=[-1\,, 1]$, in turn, at every position $x$ along column $y$, multiplying each masked element by the mask entry, and summing the result. In the discrete setting, this replicates a common signal processing calculation: \textit{convolutions}. That is, directional derivatives (in any direction) can be approximated by modifying the kernel, $k$, above. 

More specifically, given an image, $E$, of size $(m,n)$ and kernel, $k$, we can define the spatial convolution ($\otimes$) as follows: 

\begin{equation}
(E \otimes k)(x,y) = \sum_{i=0}^{m-1} \sum_{j=0}^{n-1} E(x+i,y+j)k(i,j) 
\label{eq:conv2}
\end{equation}

\begin{figure}[h!]
\centering
\includegraphics[width=0.8\linewidth]{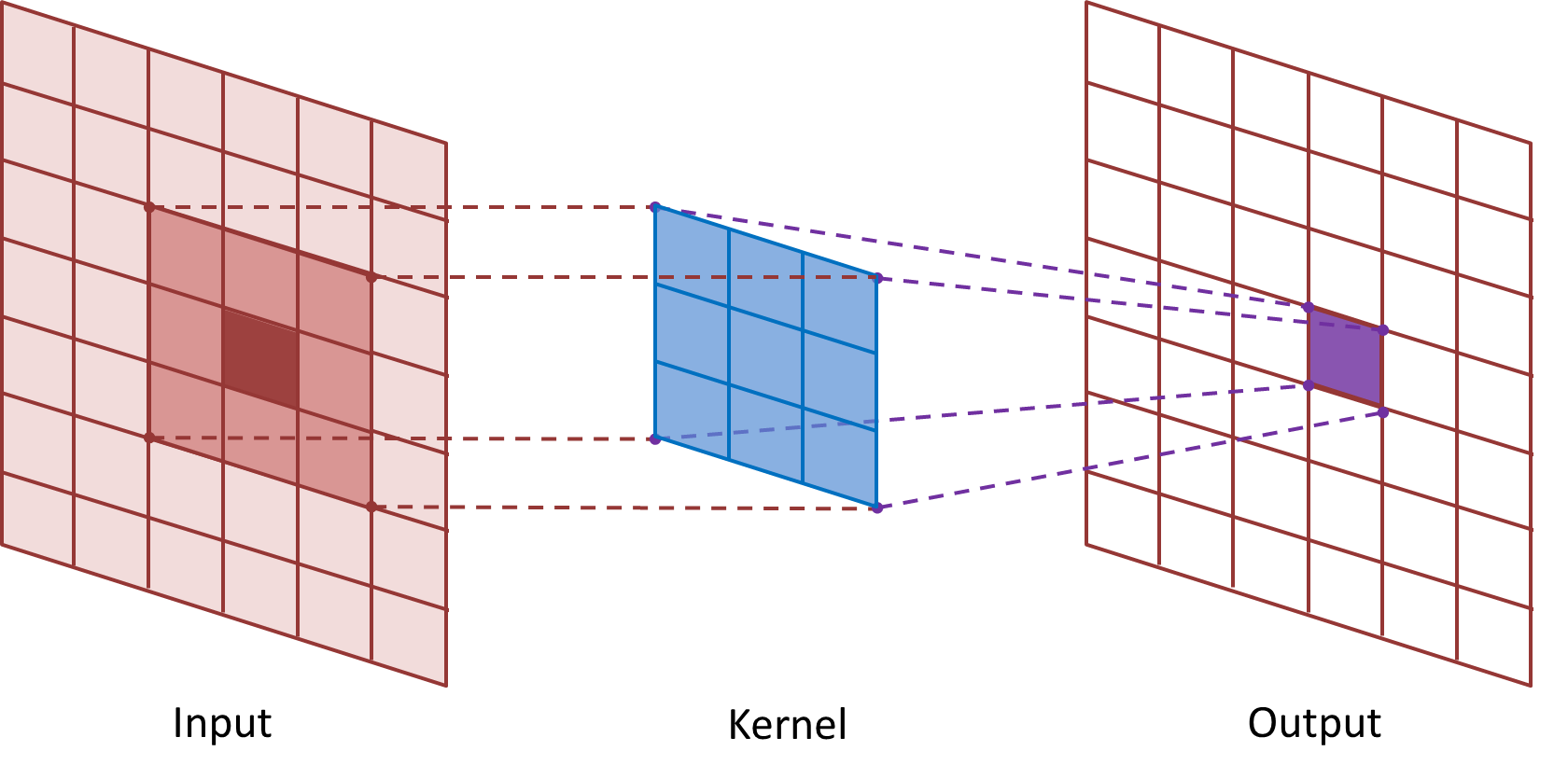}
\caption{An illustration of an input image convolved with a $3\time 3$ kernel to produce an output at the center of the kernel.}
\label{fig:conv}
\end{figure}

Figure~\ref{fig:conv} gives a visual outline of how a convolution of an image with a $3\times 3$ kernel might be performed. The kernel is applied as a sliding window that spans all possible areas of the input image, at each point the element-wise multiplication is computed and summed, which is stored in the output at the location highlighted in purple.

Now, in order to approximate the partial derivative of $E$ along the $x$ axis ($E_x$) for example, we can let:
\[ k= \frac{1}{2}
\left[ \begin{array}{ccc}
   -1 & 1\\
   -1 & 1
\end{array} \right]
\]
To find $E_y$, we use the transpose of the previous kernel. In the next section we detail some of most common kernels that have been developed over the years for the task of edge detection.

\section{Convolutions} 

Different variations of these kernels have been studied over the years to accomplish various tasks in machine vision \cite{edges}. In this section we explore three pervasive kernels, which were developed in the context of computing image derivatives. Each kernel structure has various benefits and disadvantages associated to it.

\begin{figure}[h!]
\centering
\includegraphics[width=0.8\linewidth]{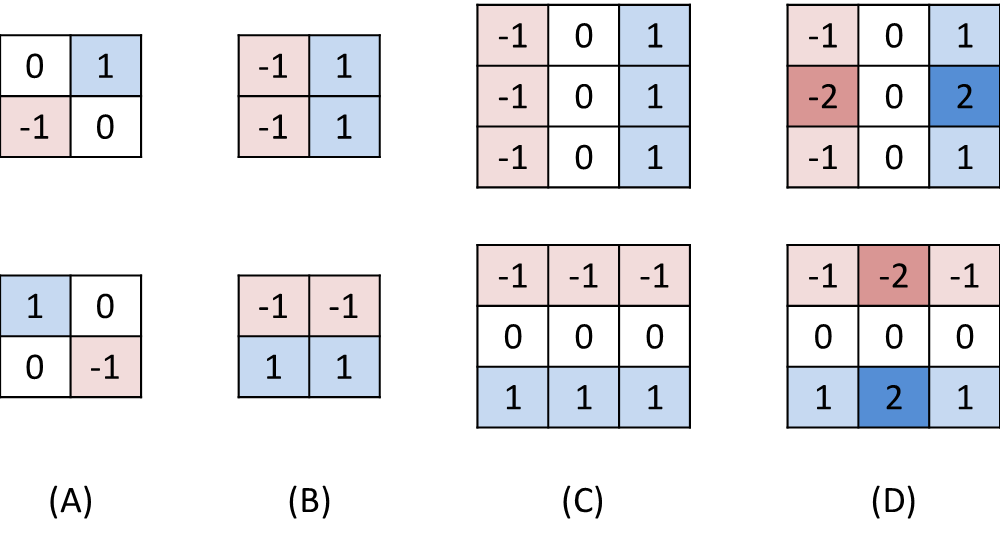}
\caption{A visual description of 2D image edge detection kernels. First and second dimensions are the top and bottom rows respectively. The different kernels are: $2\times 2$ Roberts (A), $2\times 2$ Prewitt (B), $3\times 3$ Prewitt (C), and $3\times 3$ Sobel (D).} 
\label{fig:kernels}
\end{figure}

Figure~\ref{fig:kernels} shows a visual example of the four different types of kernels that have been developed for edge detection through first derivatives. 

\subsection{Roberts}
The Roberts kernel (\ref{fig:kernels}A) uses diagonal edge gradients and is susceptible to fluctuations. This method gives no information about edge orientation and is actually computing the gradient at the center of a group of 4-neighbor pixels. Therefore, the resulting coordinate system of the output derivatives is 45 degrees away from the original Cartesian coordinate system.

\subsection{Prewitt}

The Prewitt $2\times 2$ kernel (\ref{fig:kernels}B) tries to establish a Cartesian coordinate system by defining two derivates that follow the $x$  and $y$ axes. However, there is still a slight ambiguity with this kernel. Since the kernel is of size $2\times 2$ (like Roberts), the gradient is computed at the center of the 4 pixels. Therefore, we do not know precisely \textit{where} the result pixel should be stored (ie. it is between two pixels).

Conversely, the Prewitt $3\times 3$ kernel (\ref{fig:kernels}C) addresses this issue through a well defined center (ie. at the middle pixel). Due to the nature of the kernel, it is very simple to implement but also very sensitive to noise fluctuations.

\subsection{Sobel}
Finally, the Sobel $3\times 3$ kernel (\ref{fig:kernels}D) is very similar to Prewitt kernels in that it has a well defined center.  However, it applies twice the weight around the center pixels. This makes the Sobel kernel more sensitive to diagonal edges than the previous Prewitt operators.

\section{Accelerating Convolutions}
In this section, we will present a method to speed up the computation of convolutions in a high-performance computing language, Julia \cite{julia}. We  utilize a key Julia feature of multiple dispatch, in which functions can be defined and overloaded for different combinations of argument types. Additionally, we extend our convolution algorithm to handle arbitrary $n-$dimensional convolutions, which is not supported by the current Julia implementation. $N-$ dimensional convolutions are implemented through automatic code generation of higher order convolution functions using Cartesian.jl \cite{cartesian}.

\subsection{Optimization for Vision}
Applications of convolutions in machine vision, image processing, and deep learning are unique in the sense that they utilize relatively small kernel sizes through convolution. Previous sections provided examples of computing first order derivatives with $2\times 2$ and $3 \times 3$ kernels, each containing a total of 4 and 9 pixels respectively. Such kernels are many orders of magnitude smaller than the size of the image. For example, today, it is standard for even the low-end mobile devices to possess cameras which capture 1 megapixel ($10^6$ pixels) quality images. Thus, a $3\times 3$ Sobel kernel is $\frac{10^6}{9}\approx 10^5$ times smaller than a 1 megapixel image. 

Convolutions in deep learning applications exhibit similar properties.  That is, CNNs are used in a variety of machine learning tasks, including character recognition \cite{mnist}, autonomous vehicle control \cite{nvidia}, and medical drug delivery \cite{drug}. The convolution kernels leveraged in these networks (as described above for image processing) are relatively very small, ranging from ~$2\times 2$ to ~$5\times 5$.

\subsection{Convolutions in Julia}

Historically, convolutions have been conceptualized many different ways. Although Equation~\ref{eq:conv2} gives us one representation of convolutions by computing two summations, there have also been other representations of convolutions built upon notions developed in fundamental signal processing. For example, the \textit{Convolutional Theorem} \cite{convolutional_theorem} states that convolutions in the time domain are equivalent to element-wise multiplication in the frequency domain, and vice versa. This theorem can be summarized as follows: 

\begin{equation}
E \otimes k = \mathcal{F}^{-1} \big(\mathcal{F}(E) * \mathcal{F}(k)\big)
\end{equation}  

where $\mathcal{F}$ is the Fast Fourier Transformation (FFT), and $\mathcal{F}^{-1}$ is the Inverse Fast Fourier Transformation (IFFT). In other words, the image is converted into the frequency domain (using FFT), multiply, and convert back to the time domain (using the IFFT). Complexity analysis of this algorithm reveals that it exhibits $O(n \log n)$ asymptotic behavior (where $E$ and $k$ have $n$ samples each). Theoretically, this indicates the FFT approach to computing convolutions would out-perform the direct summation method outlined in equation~\ref{eq:conv2}, which has complexity $O(n^2)$. For these reasons: (1) compact implementation and (2) lower computational complexity, this approach was chosen for the existing convolution implementation in the Julia software package. 

However, while the FFT/IFFT approach to computing convolutions does exhibit low asymptotic convergence, in many practical machine learning and machine vision scenarios, a far more efficient approach can be provided. This is due to the fact that, when using kernels that are orders of magnitude smaller than the raw signal (i.e., image), the overhead of the FFT/IFFT overshadows any efficiency advantages gained through the transformation to the frequency domain. Thus, to take advantage of the small kernels leveraged by machine vision and machine learning, this study implements a new direct method of convolution computation and evaluates performance results in these settings. 

The goal of our implementation is to develop a novel, hybrid implementation, which exploits the small kernel sizes where possible for low complexity computation, and leverages the existing FFT approach only when dealing with large kernel sizes (which does not occur frequently). Furthermore, unlike many other comparable high performance computing languages, Julia does not have built-in functionality to compute n-dimensional convolutions. We will address this gap by extending our algorithms into an arbitrary n-dimensional implementation to handle a much larger range of applications. 

\begin{figure}[h!]
\includegraphics[width=0.8\linewidth]{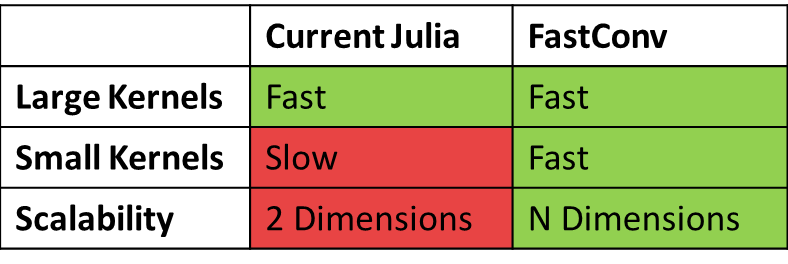}
\centering
\caption{A conceptual outline of the benefits of the implementation developed in this paper (FastConv), to the current implementation that exists within Julia.}
\label{fig:goals}
\end{figure}

This study implements and evaluates an algorithm, \textbf{FastConv}, in Julia that aims to accomplish the goals outlined in Figure~\ref{fig:goals}.

\subsection{FastConv}
Due to the high overhead in converting a signal into the frequency domain, the algorithm starts by evaluating the kernel size. If the kernel size is relatively small, the convolution will be performed in a direct manner (ie. via equation~\ref{eq:conv2} for 2D); otherwise the existing FFT method will be used. Kernel size checking incurs a small increase in computation, but avoids requiring the user to assess which method to use, and is dwarfed by the computations required for both the direct and the FFT methods. 

FastConv \cite{fastconv} employs use of existing base Julia features (such as multiple dispatch, efficient memory allocations, and macros to generate multi-dimensional code). The package, which has been added to the official Julia Package Manager, achieves over an order of magnitude speedup for small kernel applications (such as machine learning and machine vision), e.g., when convolving a standard sized image with a $3\times3$ kernel. By allocating the memory in a helper function, the algorithm also improves efficiency by reusing memory within the actual convolution function. Additionally, we use \texttt{@inbounds} to eliminate array bounds checking, further increasing performance speedups. 

Our implementation covers a wide variety of signal types (real, complex, boolean, irrational, unsigned integers, etc) through native Julia multiple dispatch. Furthermore, we utilize the Cartesian package to create auto generative code that can compute convolutions in any dimension. The process by the compiler is as follows:

\begin{enumerate}
\item The dimensionality of the inputs is identified (let's say both inputs are of dimension $k$)
\item The $k$-dimensional convolution code is generated (if it has never been before)
\item The inputs are processed by the auto-generated code. 
\end{enumerate}

\subsection{Benchmarking}
\begin{figure}[h!]
\includegraphics[width=\linewidth]{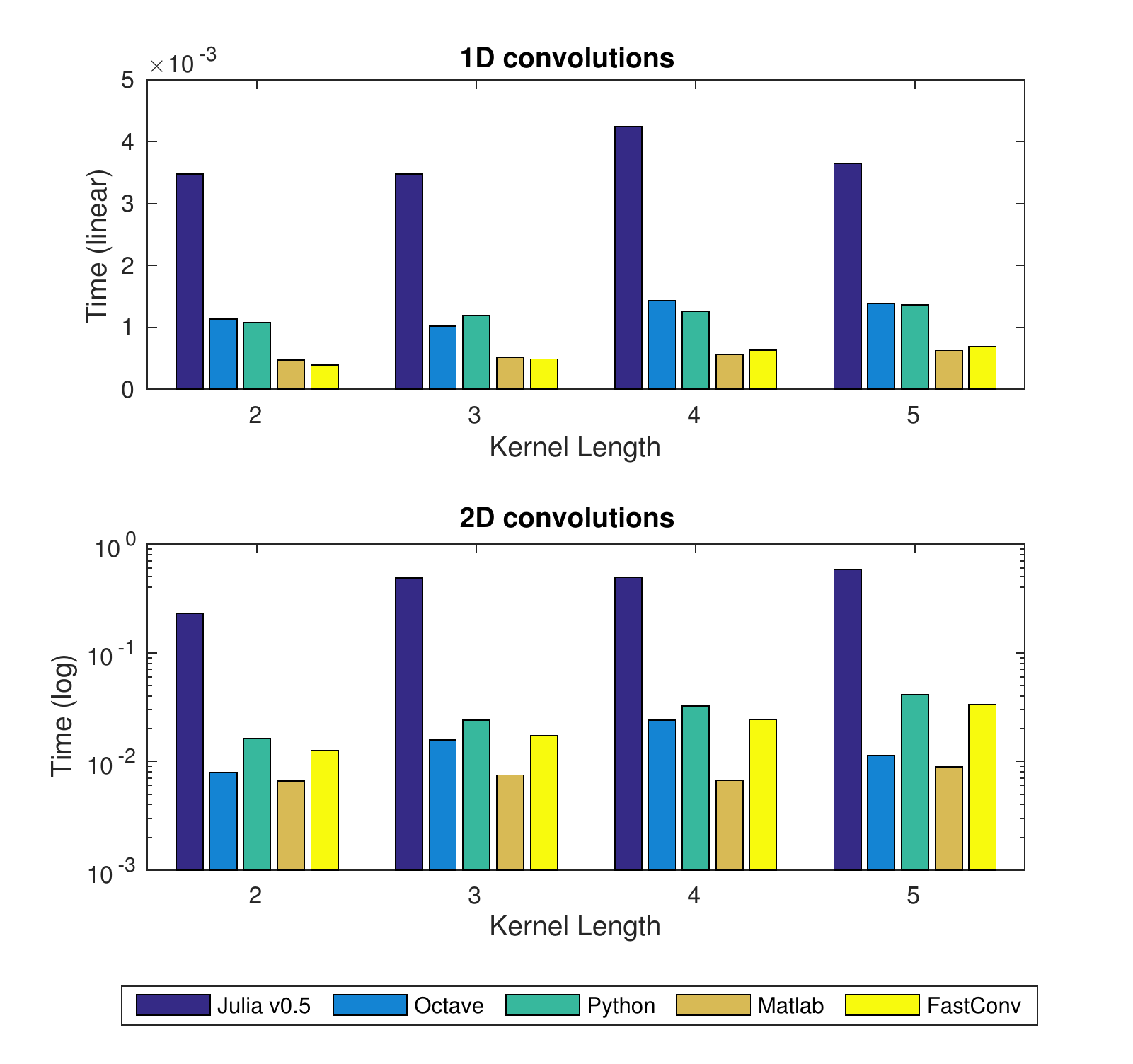}
\centering
\caption{Time taken to compute $E\otimes k$ across three different computing platforms (Julia, Matlab, and Python). For each language the built in convolution function was used. Note that Julia does not have base support for convolutions of arbitrary dimension (above 2D).}
\label{fig:benchmark_stats}
\end{figure}

To compare the performance of select programming languages at computing convolutions, a simple benchmark, designed to represent machine vision applications, was conducted.  A random 1MP image was convolved with 4 different sized kernels. Figure~\ref{fig:benchmark_stats} shows the time taken to convolve these two signals using the standard convolution function in three different computing platforms (\texttt{conv} in Julia, Matlab, Octave or \texttt{numpy.convolve} in Python). 

The number of elements present in the input image was maintained for each test (ie. only the language and kernel size was varied). Since Julia does not support for convolutions of dimension three and above, performance was compared to Julia only for 1D and 2D inputs.

\begin{figure}[h!]
\includegraphics[width=0.8\linewidth]{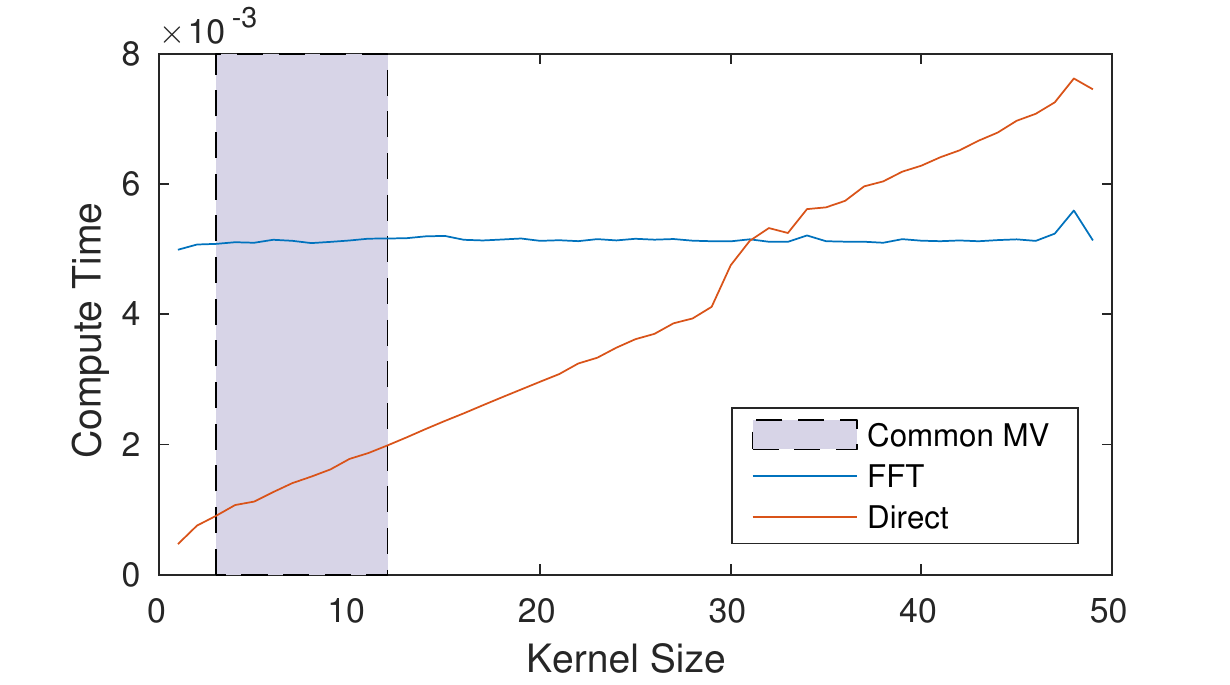}
\centering
\caption{Convolution computation time for both the direct (red) and FFT (blue) method across varying kernel sizes. FastConv uses a threshold on the kernel size to determine when compute according to either method. Frequently used machine vision kernel sizes (as described in the main text) are highlighted in blue background. }
\label{fig:scaling}
\end{figure}

FastConv achieved well over an order of magnitude speedup over the convolution function in the current Julia version (v0.5), and performed comparably to Python, Matlab and Octave implementations for these kernels. Additionally, we evaluated our algorithm against the FFT based approach for kernels of larger size. Figure~\ref{fig:scaling} plots the time taken to compute the convolution across varying kernel sizes for both the pre-exsiting Julia implementation and FastConv. 

For the majority of machine vision (MV) applications (shaded region), FastConv significantly outperforms the FFT implementation. As the kernel size increases (kernel size $\approx 30$) it becomes more efficient to use an FFT based convolution algorithm. Therefore, in the published package of FastConv we utilize the best of both worlds. When below this threshold (determined experimentally), which is common in the realm of machine vision, the algorithm automatically uses the direct method of computing convolutions. However, for large kernel sizes, the algorithm switches to use the FFT based implementation.

\section{Time to Contact}

In order to test how FastConv might be used in the context of machine vision, we also implemented the time to contact algorithm in Julia. This section will briefly outline the details of the algorithm we follow to determine precisely the focus of expansion (FOE) and resulting time to contact (at the FOE) \cite{ttc}.

\subsection{Formulation}
\begin{figure}[h!]
\centering
\includegraphics[width=0.8\linewidth]{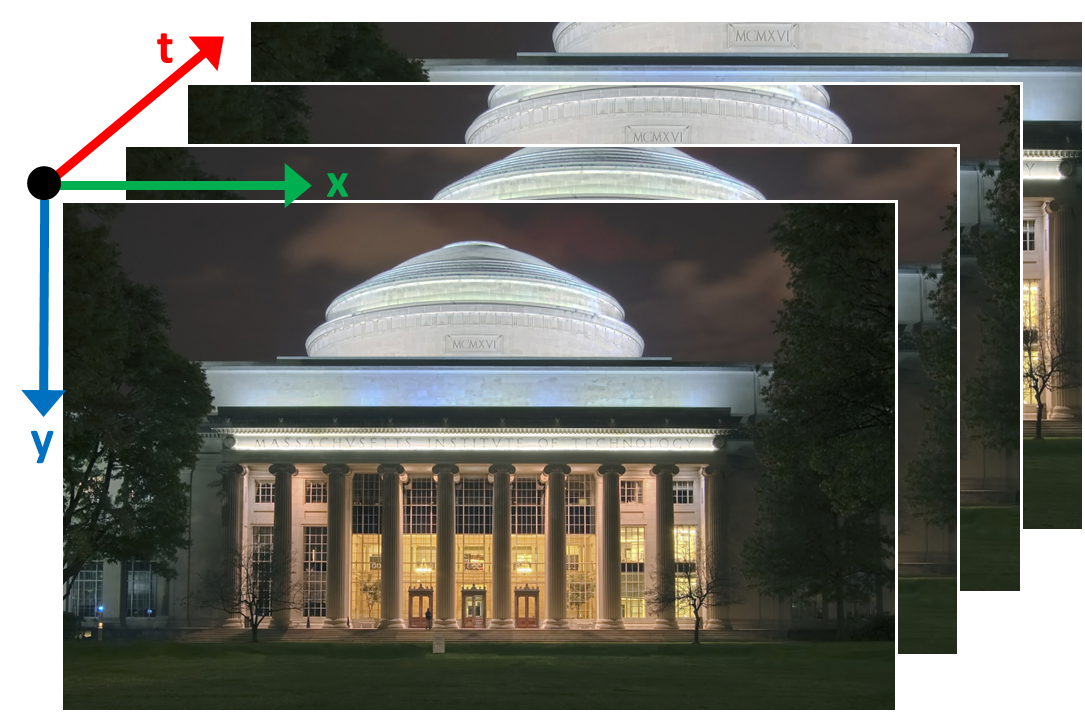}
\caption{A sequence of four images of the MIT dome, zooming into the dome as time progresses. Images vary spatially according to $x$ and $y$, while time is represented along the temporal (ie. $t$) axis.} 
\label{fig:Eaxis}
\end{figure}

We can define the brightness of an image sequence over time as $E(x,y,t)$, where $x$ and $y$ are Cartesian coordinates across the image and $t$ is the temporal operator. For example, $E(x,y,0)$ is the image $E(x,y)$ at time $t=0$. Figure~\ref{fig:Eaxis}, illustrates how this sequence can be visually represented as ``stacking'' images on top of one another to capture the temporal variability. Studying how a certain pixel, or group of pixels, vary over a temporal slice of the dataset can give us insights into optical flow and time to contact. 

To start, assume that as a point moves through the scene, its brightness value (in $E$) does not vary significantly. \cite{flow} Therefore according to the chain rule for differential calculus, we get:
\begin{align*}
\frac{d}{dt} E(x,y,t) &= 0\\
\frac{\partial E}{\partial x}\frac{dx}{dt} + \frac{\partial E}{\partial y}\frac{dy}{dt} + \frac{\partial E}{\partial t} &= 0\\
E_x\frac{dx}{dt} + E_y\frac{dy}{dt} + E_t &= 0
\end{align*}
\begin{equation}
u\, E_x + v\,E_y + E_t = 0
\label{eq:opticalflow}
\end{equation}

where $E_x$ and $E_y$ are the derivatives of the image in the $x$ and $y$ directions respectively. 

\begin{figure}[h!]
\centering
\includegraphics[width=1\linewidth]{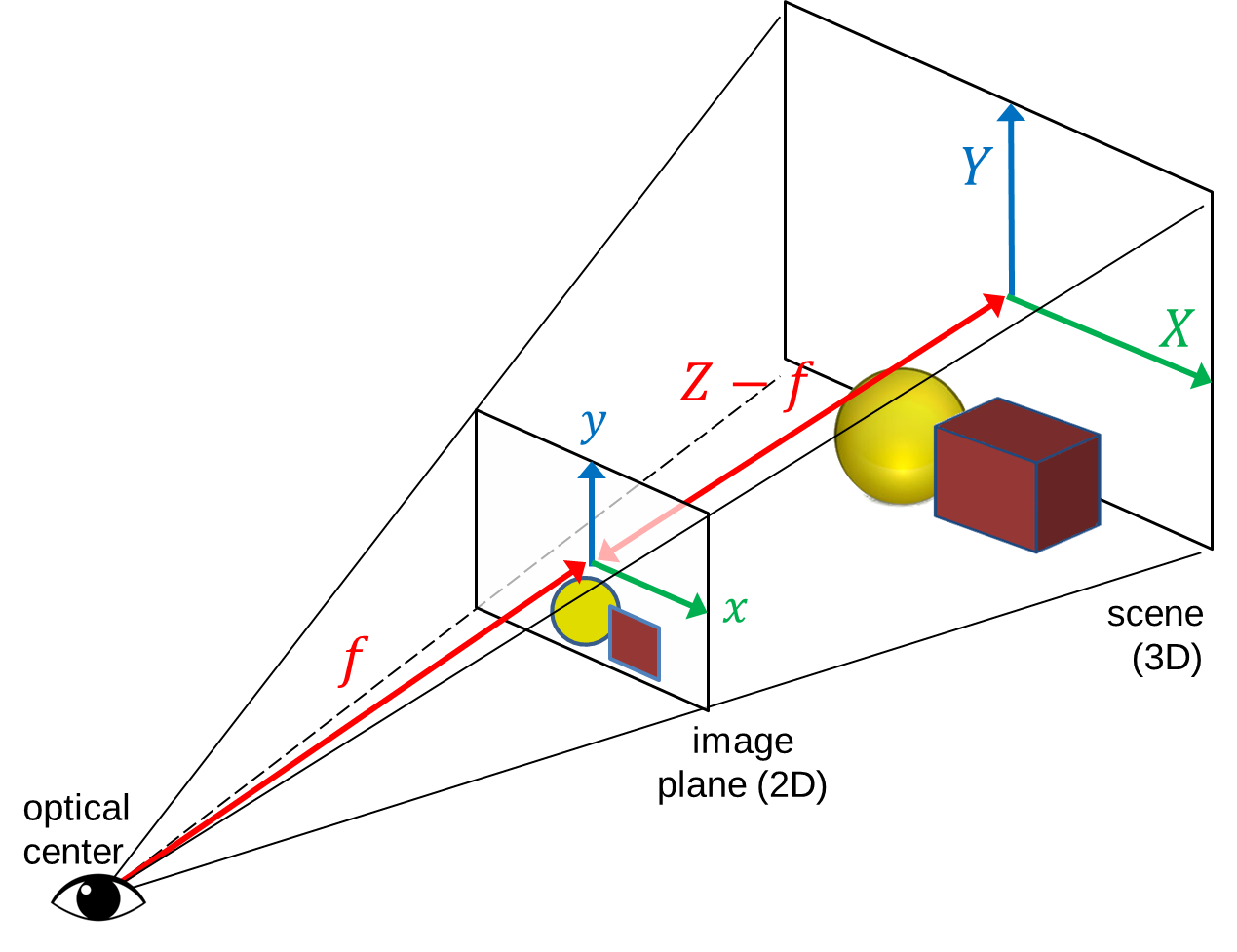}
\caption{Diagram illustrating how 3D object in the environment can be projected into a 2D image plane. From the point of view of the center of projection (eye), points in the image plane and in the environment form similar triangles that we use to define the perspective projection equations.} 
\label{fig:projection}
\end{figure}

In order to gain a better understanding of the way objects are situated in our environment compared to within our images, we can establish the following relationship of perspective projection. Given a optical center (or center of projection) with focal length $f$, we can use the rule of similar triangles to conclude: 

\begin{align}
\frac{x}{f} &= \frac{X}{Z},  &&&  \frac{y}{f} &= \frac{Y}{Z}
\label{eq:pp_start}
\end{align}

where $X,Y,Z$ are the positions of a point in space, and $x,y$ is the position of that same point projected onto the image plane. Figure~\ref{fig:projection} gives a visual representation of a scene (3D) being projected on to an image plane (2D), and the proportional relationship between any point in the scene to its corresponding point in the image (ie. two equations above). 

Differentiating equations (\ref{eq:pp_start}) with respect to time yields: 
\begin{align}
\frac{u}{f} &= \frac{1}{Z}\left( U-\frac{x}{f}W \right) ,&&& \frac{v}{f} &= \frac{1}{Z}\left( V-\frac{y}{f}W \right)
\label{eq:ppcombined}
\end{align}
where $(U,\,V,\,W)$ are the time derivatives of $(X,\,Y,\,Z)$ respectively.

This can be rewritten in terms of the focus of expansion $(x_0,y_0)$ : 
\begin{align*}
u &= -\frac{W}{Z}(x-x_0) ,&&& v&=-\frac{W}{Z}(y-{}y_0)
\end{align*}

In other words, define the focus of expansion and time to contact as follows: 
\begin{align}
x_0 = -\frac{A}{C} && y_0 = -\frac{B}{C} && TTC = \frac{1}{C}  
\label{eq:final_TTC}
\end{align}

where $A=f\frac{U}{Z}$, $B=f\frac{V}{Z}$, and $C=\frac{W}{Z}$. Finally, equations (\ref{eq:ppcombined}) can be substituted into the original formula for constant optical flow, equation (\ref{eq:opticalflow}). After simplifying: 

\begin{align*}
(A+C x)E_x + (B+C y)E_y + E_t &= 0\\
AE_x + BE_y + C(xE_x+yE_y) + E_t &= 0\\
AE_x + BE_y + CG + E_t &= 0\\
\end{align*}

with $G=xE_x+yE_y$ representing a measure of the ``radial gradient''. The least squares optimization problem is formulated as follows: 

\begin{align*}
\min_{A,B,C} \iint(AE_x + BE_y + CG + E_t)^2 \,dx\,dy
\end{align*}

Taking derivatives with respect to $A$, $B$ and $C$ individually and setting equal to zero, a system of three linearly independent equations can be obtained and written in matrix form as:

{\footnotesize
\[ \left[ \begin{array}{ccc}
   \iint E_x^2 & \iint E_x E_y & \iint GE_x\\
   \iint E_xE_y & \iint E_y^2 & \iint GE_y\\
   \iint G E_x & \iint G E_y & \iint G^2\\
\end{array} \right] \colvec{3}{A}{B}{C} = \colvec{3}{-\iint E_x E_t}{-\iint E_y E_t}{-\iint GE_t}\\
\]}

Finding the inverse of the $3\times3$ matrix on the left enables solving for the vector of interest: $[A\, B\, C]^T$. Once these values are determined, the TTC and FOE can be computed by plugging into the respective equation (\ref{eq:final_TTC}) and finding $[x_0,\, y_0,\, TTC]$.

\subsection{Results}
The TTC and FOE determination algorithm described in the previous subsection was also implemented and integrated into the official Package Manager built into Julia. We tested our implementation across a variety of synthetic videos that were generated by progressively ``zooming'' into random points on the image, each time performing a cubic interpolation maintain image size. 

\begin{figure*}[h!]
\centering
\includegraphics[width=1\linewidth]{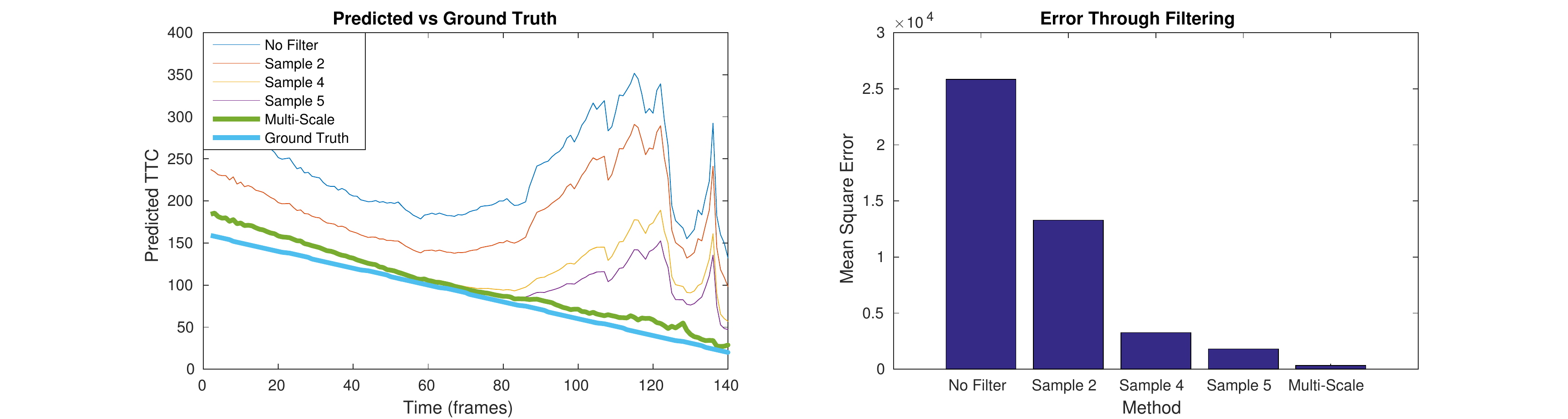}
\caption{Predicted time to contact vs the ground truth for a synthetically generated video sequence (left). Five down sampling methods are used, from none to five consecutive down samplings. The right subplot illustrates the mean squared error for each of the methods used.} 
\label{fig:ttcres}
\end{figure*}

Given any input video pointer (ie. pointer to local video, or to webcam), the algorithm moves through the data to progressively determine the time to contact at that instant, along with the location that will collide first (ie. FOE). Figure~\ref{fig:ttcres} shows the results of the TTC algorithm compared to ground truth data for a certain video sequence using 5 different filtering methods to enhance results. 

A common disadvantage of having an image with a very large number of pixels is that it becomes increasingly harder to mitigate the effects of noise and distortions. To combat this, and reduce noise we can down sample each image by filtering it with a block averaging filter, and then passing the result through a low-pass filter. This process will create a smaller image, with inevitably less noise. We can continue this further, each time removing more and more noise (but also loosing more and more image detail). The five methods shown in Figure~\ref{fig:ttcres} represent different levels of down-sampling. In other words, ``No Filter'', means the original images were used as inputs to the TTC algorithm, while ``Sample'' 2, 4, 5 represent the act of down sampling the inputs 2, 4, 5 consecutive times before even passing them to the TTC algorithm. 

We observe that filtering certainly helps up to a certain point, which may be hard to determine at first glance. However, a ``Multi-Scale'' technique can also be employed to progressively run a search over the sampling space to identify the best possible amount to down sample at every video frame. For example, we start with no sampling, compute the expected TTC, down-sample and recompute the TTC. If the estimate improved (ie. reduced) then we continue to down sample until no additional progress is made. 

This technique achieves significantly better results that ad hoc sampling of the entire video sequence, and requires no additional prior knowledge about the environment, or ground truth TTC. These results are illustrated by the green line on the left plot of Figure~\ref{fig:ttcres}, while the respective mean squared error (MSE) for each of the methods is shown on the right. Figure~\ref{fig:ttcres} illustrates the multi-scale TTC algorithm significantly outperforms any of the static scaling methods, but also requires more computation to progressive search through the sampling space. Testing our algorithm on a synthetically generated standard definition video sequence, we are able to obtain real-time performance of both the original TTC algorithm ($\approx30$Hz) as well as the multi-scale variant ($\approx10$Hz).

\section{Future Work}
We are currently capturing video data from multiple cameras mounted on a self-driving automobile in the MIT Distributed Robotics Laboratory. This data presents new challenges, such as multiple points of interest, and potentially excessive noise (snow, rain, reflections, etc), and will enable extending and evaluating our algorithms against real-world driving conditions. We are also extending our implementation to automatically detect separable kernels (i.e., matrices that are capable of being represented as the multiplication of two vectors) and compute the convolution piece-wise over each segment of the separable kernel. This would reduce the complexity of the direct algorithm from $O(n^2)$ to $O(n)$, and bring additional performance gains to our implementation. Separable kernels are a special set of kernels that appear frequently in machine vision, as (1) directional derivatives (2) Gaussian smoothing function, and (3) block averages.  

Additionally, we aim to extend our implementations for applications outside of machine vision. Specifically, we plan to extend our direct method of computing n-dimensional convolutions into the FFT method, in order to handle convolutions with large kernel sizes.

\section{Conclusion}
In this paper, we implement a fast N-dimensional convolution algorithm, optimized specifically for machine vision applications, and integrate into the high performance computing Julia platform. We show that our algorithms achieve an order of magnitude performance improvement over the existing implementation. 

Furthermore, we develop a real time implementation of time to contact and focus of expansion determination using a single stereo video sequence. We also evaluate our accelerated convolution implementation in the context of real-time TTC and FOE determination. In autonomous systems, TTC and FOE provide a low complexity method to estimate the danger of their trajectory and trigger evasive maneuvers as needed. Our accelerated, low complexity, and reduced memory implementation is especially attractive for resource-constrained embedded devices and autonomous vehicle control. 

Results are measured against synthetically generated videos and quantitatively assessed according to their mean squared error from the ground truth. Utilizing multiple scales of resolution we show that it is possible to drastically improve the overall performance of the TTC estimation. 

Finally, we packaged and published our code \cite{fastconv,ttcjulia}, for both the accelerated convolutions and the TTC algorithm, into the official Julia Package Manager, allowing our results to be leveraged on any Julia capable device with only a few lines of code.

\end{document}